\documentclass[10pt,letterpaper]{article}
\usepackage{memtrace-arxiv}
\newcommand{\shortpapertitle}{MemTrace}

\newcolumntype{L}[1]{>{\raggedright\arraybackslash}p{#1}}

\newcommand{\mtbest}[1]{\underline{#1}}
\newcommand{\mtgain}[1]{\textcolor{MTGreen}{\textbf{#1}}}
\newcommand{\mtloss}[1]{\textcolor{MTRed}{\textbf{#1}}}

\newcommand{\code}[1]{\texttt{\detokenize{#1}}}
\DeclareRobustCommand{\codepath}[1]{{\urlstyle{tt}\small\path{#1}}}

\newcommand{\TracePageBibliography}{%
  \nocite{*}%
  \bibliography{references}%
  \bibliographystyle{iclr2027_conference}%
}

\title{MemTrace: State-Consistent Memory for Long-Horizon Coding Agents}
\newcommand{\paperauthors}{%
  \mbox{\textbf{\href{https://openreview.net/profile?id=~Hongming_Xu5}{\textcolor{black}{Hongming Xu}}}\textsuperscript{1,2,3}}\quad
  \mbox{\textbf{\href{https://openreview.net/profile?id=~Le_Zhou4}{\textcolor{black}{Le Zhou}}}\textsuperscript{1,2,3}}\quad
  \mbox{\textbf{\href{https://openreview.net/profile?id=~ZhongHe_Jin1}{\textcolor{black}{ZhongHe Jin}}}\textsuperscript{1,2}}\quad
  \mbox{\textbf{\href{https://openreview.net/profile?id=~Xiang_Zhang65}{\textcolor{black}{Xiang Zhang}}}\textsuperscript{1,2}}\\[0.3em]
  \mbox{\textbf{\href{https://openreview.net/profile?id=~Bo_Tang3}{\textcolor{black}{Bo Tang}}}\textsuperscript{3}}\quad
  \mbox{\textbf{\href{https://openreview.net/profile?id=~Zhiyu_li2}{\textcolor{black}{Zhiyu Li}}}\textsuperscript{3}}\quad
  \mbox{\textbf{\href{https://openreview.net/profile?id=~Xuanhe_Zhou1}{\textcolor{black}{Xuanhe Zhou}}}\textsuperscript{1,2,\ensuremath{\dagger}}}\quad
  \mbox{\textbf{\href{https://openreview.net/profile?id=~Juncheng_Zhang4}{\textcolor{black}{Juncheng Zhang}}}\textsuperscript{3,\ensuremath{\dagger}}}
}
\author{%
  \paperauthors\\[0.5em]
  {\small\textsuperscript{1}Shanghai Jiao Tong University\quad
  \textsuperscript{2}Theseus Lab\quad
  \textsuperscript{3}\href{https://memtensor.cn}{\textcolor{black}{MemTensor (Shanghai) Technology Co., Ltd.}}\\[0.15em]
  \textsuperscript{\ensuremath{\dagger}}Corresponding authors.}%
}

\hypersetup{
  pdftitle={MemTrace: State-Consistent Memory for Long-Horizon Coding Agents},
  pdfauthor={Hongming Xu, Le Zhou, ZhongHe Jin, Xiang Zhang, Bo Tang, Zhiyu Li, Xuanhe Zhou, Juncheng Zhang}
}

\renewcommand{\preprintlogos}{%
  \raisebox{-0.5\height}{\includegraphics[width=0.25\linewidth]{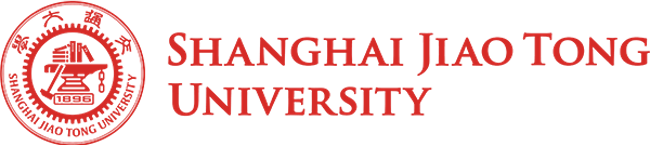}}\hfill
  \raisebox{-0.5\height}{\includegraphics[width=0.25\linewidth]{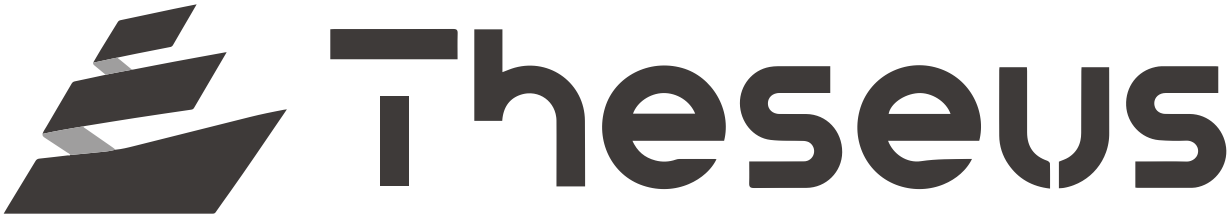}}\hfill
  \raisebox{-0.5\height}{\includegraphics[width=0.25\linewidth]{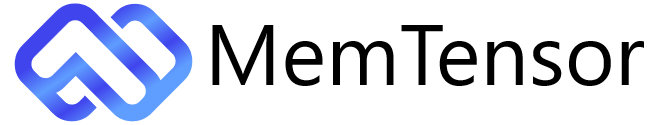}}%
}

\begin{document}
\maketitle

\begin{abstract}
As coding agents take on long-horizon software evolution tasks spanning multiple files and stages, longer execution trajectories introduce two coupled challenges: (1) accumulated histories strain context budgets, and (2) repository changes can invalidate earlier execution evidence. Existing approaches address these challenges through techniques like larger context windows, compression, retrieval, or repository representations, but often fail to reconstruct a consistent task state after a context refresh or verify whether recalled evidence remains valid. Thus, we introduce \textbf{MemTrace}, a provenance-aware memory system that preserves execution history and aligns its reuse with the evolving task (e.g., iterative cross-file repair) and repository state. MemTrace stores history as immutable \emph{Memory Traces} anchored to key information (e.g., files, symbols, tests), and organizes their execution order and dependencies in a \emph{Memory Trace Graph}. When context is constrained, working memory retains only compact \emph{Memory Anchors}, from which the agent can reconstruct the latest execution state and locate evidence relevant to its next action. Before restoring historical evidence, MemTrace checks its validity against the current repository state and retrieves only what the next action requires. Across three complementary long-horizon coding benchmarks, MemTrace consistently outperforms all fully evaluated baselines under the same backbone and harness, \textit{improving DeepSWE pass@1 by 21.2 points, SWE-EVO Resolved Rate by 4.4 points, and SWE-Milestone Score by 17.8 points under Codex CLI.} 
\par\vspace{0.4em}
\noindent{\small\textbf{GitHub: }\href{https://github.com/theseus-labs-rsi/MemTrace}{https://github.com/theseus-labs-rsi/MemTrace}}\par
\end{abstract}

\section{Introduction}
\label{sec:intro}
Coding agents are moving from isolated bug fixing toward cross-file,
multi-stage software maintenance and development, which requires iterative
code inspection, coordinated edits, testing, and
revision~\citep{jimenez2024swebench,thai2025sweevo,huang2026deepswe}.
As execution lengthens, growing interaction history increases context pressure,
while ongoing repository changes can make earlier observations, test outcomes,
and decisions no longer applicable to the current state.
This challenge already appears in deployed coding agents: systems such as
Codex CLI~\citep{bolin2026codex} use context compaction to continue
long-running tasks across context boundaries.
Such mechanisms preserve execution continuity, but continuity alone does not
determine whether recovered evidence still applies after the repository has
changed.
Long-horizon execution therefore requires the agent to maintain an accurate
representation of task progress: what has been completed, what remains
unresolved, and which prior evidence still applies.
\emph{Task-state drift} arises when this representation no longer matches
the current task and repository state.
The problem becomes particularly important after a context refresh, when the
agent must reconstruct prior progress and recover evidence that remains
applicable before continuing the unfinished task.

Existing approaches mitigate different parts of this problem.
Longer context windows accommodate more interaction history, while prompt
compression condenses earlier context and observation masking selectively
removes prior observations and tool outputs.
External-memory systems offload historical information for later
retrieval~\citep{packer2023memgpt,lee2024readagent,chen2023memwalker},
while repository graphs represent files, symbols, and dependencies for code
localization~\citep{liu2024codexgraph,ouyang2025repograph}.
These approaches improve history access and code localization, but access
alone does not determine which historical evidence still supports the task
state being resumed.

\begin{figure}[!t]
    \vspace{-1em}
    \centering
    \includegraphics[
    width=\linewidth,
    trim=20pt 10pt 20pt 10pt,
    clip
]{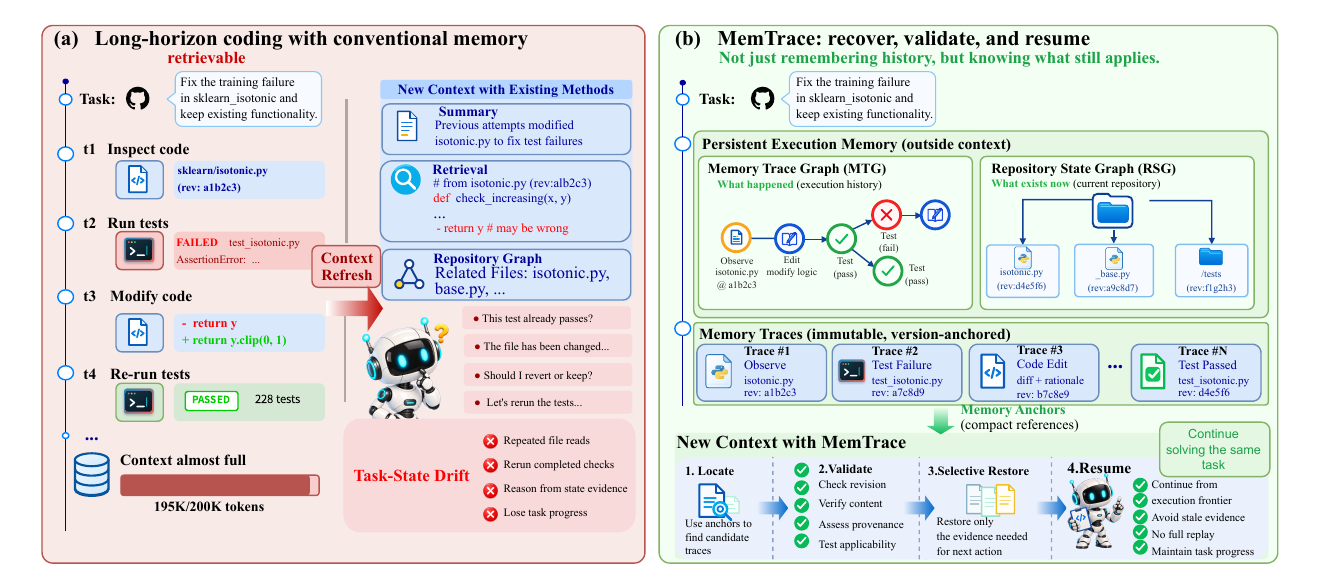}
\caption{
\textbf{Task-state recovery across context refreshes.}
    (a) Existing approaches use summaries, retrieved history, or repository structure, but may lose task progress or reuse stale evidence.
    (b) MemTrace reconstructs the latest execution state, validates historical evidence against the current repository, and restores only what is needed to resume the task.
}
    \label{fig:intro_overview}
    \vspace{-1em}
\end{figure}


The remaining challenge is to recover the current task state, not merely
historical content. As illustrated in Figure~\ref{fig:intro_overview}(a), a refreshed context
may retrieve an earlier test failure even after a repair and a passing rerun.
Without the relation to these later events, the failure may be mistaken for
an unresolved issue. We call this mismatch between recovered history and the current task and
repository state \emph{state-memory misalignment}. Addressing it requires reconstructing the current task state and
identifying which prior evidence remains valid in the current repository
and relevant to the next action.

To address this gap, we introduce \textbf{MemTrace}, a provenance-aware
memory system that casts long-horizon coding memory as a state-alignment
problem rather than mere history retention.
Completed execution is sealed into immutable Memory Traces with
explicit repository bindings and organized through two complementary views.
A Memory Trace Graph (MTG) encodes \emph{depend}, \emph{verify},
\emph{correct}, and \emph{supersede} relations, enabling relation-aware
traversal from the current execution frontier to evidence that establishes,
revises, or invalidates its state.
A Repository State Graph (RSG) maintains the current files, symbols,
tests, and their structural relations. Trace-to-entity bindings project
candidate traces onto this evolving state to determine whether their
referenced evidence remains applicable.
Under context pressure, trace bodies are offloaded while Memory Anchors preserve addressability.
After a context refresh, MemTrace reconstructs the frontier through the MTG,
checks candidate evidence against the RSG, filters stale or superseded
traces, and loads only the validated evidence required for the next action
back into working context
(Figure~\ref{fig:intro_overview}(b)).

To assess MemTrace's effectiveness in long-horizon coding, we evaluate it
on three complementary benchmarks covering repository-level engineering,
multi-step software evolution, and continuous milestone-based development:
DeepSWE, SWE-EVO, and SWE-Milestone.
MemTrace consistently outperforms fully evaluated baselines across agent
harnesses, while achieving a 46.4\% overall score on SWE-Milestone under Codex.
Ablations show that persistence alone is insufficient: both structured
execution memory and repository-state grounding are important for maintaining
state-consistent memory over long-horizon execution.

\noindent\textbf{Contributions.}
This paper makes the following contributions.\\
(1) We introduce a state-bound execution-memory representation for
long-horizon tasks that preserves historical evidence with its original
repository bindings and records how later execution supports, revises, or
supersedes earlier evidence.\\
(2) We present MemTrace, which links the MTG and RSG through shared
repository references to reconstruct task state and selectively restore
applicable historical evidence after a context refresh.\\
(3) On three long-horizon coding benchmarks, MemTrace consistently outperforms
the evaluated baselines under matched backbone and agent harness, while
ablations confirm the complementary roles of persistent storage, structured
execution memory, and repository-state grounding. 
\section{Related Work}
\label{sec:related_work}

\providecommand{\mtfull}{\ensuremath{\checkmark}}
\providecommand{\mtpart}{\ensuremath{\triangle}}
\providecommand{\mtnone}{\ensuremath{\times}}

\begin{table}[!b]
\vspace{-1em}
\centering
\caption{Comparison of representative methods discussed in Related Work.
\mtfull, \mtpart, and \mtnone{} denote explicit, partial, and no support
under the criteria below.}
\label{tab:comparison}
\begingroup
\footnotesize
\setlength{\tabcolsep}{2pt}
\renewcommand{\arraystretch}{1.13}
\begin{tabular}{@{}
  >{\raggedright\arraybackslash}p{0.22\linewidth}
  *{5}{>{\centering\arraybackslash}p{0.14\linewidth}}@{}}
\toprule
\textbf{Method}
& \shortstack{\textbf{Context}\\\textbf{management}}
& \multicolumn{2}{c}{\textbf{Execution state}}
& \multicolumn{2}{c}{\textbf{Repository alignment}} \\
\cmidrule(lr){2-2}\cmidrule(lr){3-4}\cmidrule(l){5-6}
& \shortstack{Recoverable\\history}
& \shortstack{Structured\\task state}
& \shortstack{Revision\\history}
& \shortstack{Code\\graph}
& \shortstack{Repo-aware\\evidence reuse} \\
\midrule
Observation Masking & \mtnone & \mtnone & \mtnone & \mtnone & \mtnone \\
MemGPT             & \mtfull & \mtnone & \mtnone & \mtnone & \mtnone \\
Context as a Tool  & \mtpart & \mtpart & \mtnone & \mtnone & \mtnone \\
Self-GC            & \mtfull & \mtnone & \mtnone & \mtnone & \mtnone \\
Scroll             & \mtfull & \mtpart & \mtpart & \mtnone & \mtnone \\
\midrule
Zep                & \mtfull & \mtnone & \mtfull & \mtnone & \mtnone \\
MAGE               & \mtfull & \mtfull & \mtfull & \mtnone & \mtnone \\
Ledger             & \mtnone & \mtfull & \mtpart & \mtnone & \mtfull \\
\midrule
RepoGraph          & \mtnone & \mtnone & \mtnone & \mtfull & \mtnone \\
\midrule
\textbf{MemTrace}  & \mtfull & \mtfull & \mtfull
                   & \mtfull & \mtfull \\
\bottomrule
\end{tabular}
\par\smallskip
\begin{minipage}{\linewidth}
\footnotesize
\emph{Note.} History: \mtfull{} original records remain recoverable after
leaving context; \mtpart{} only summaries remain. Task state: \mtpart{}
implicit progress. Revisions: \mtpart{} event order or staleness without
explicit correction. Repo-aware reuse checks current code; Ledger requires
still-visible outputs.
\end{minipage}
\endgroup
\end{table}

\paragraph{Context management for long-horizon agents.}
Long-horizon agents must manage growing interaction histories within a finite context window. LLMLingua and LongLLMLingua compress prompts~\citep{jiang2023llmlingua,jiang2024longllmlingua}, while observation masking and SWE-Pruner reduce the tool content retained in context~\citep{lindenbauer2025complexitytrap,wang2026swepruner}. MemGPT manages information across working and external memory~\citep{packer2023memgpt}; ReadAgent, MemWalker, and RAPTOR use summaries and retrieval to access long documents~\citep{lee2024readagent,chen2023memwalker,sarthi2024raptor}. For agent trajectories, Context as a Tool and Context-Folding organize history around tasks or subtrajectories~\citep{liu2025contexttool,sun2025contextfolding}, while Self-GC and Scroll retain recoverable history outside the active context~\citep{hao2026selfgc,lin2026scroll}. These approaches make longer histories manageable. However, recovering historical content does not establish whether it still supports the task after later edits or tests. MemTrace couples history retrieval with task-state recovery, using relations among execution records to determine what has been completed and what remains unresolved.

\paragraph{Agent memory and execution-state management.}
Agent-memory systems retain experiences and reflections for later decisions~\citep{park2023generative,shinn2023reflexion}. Mem0 updates salient memories, A-MEM links related memory notes, and Zep tracks the temporal validity of facts while retaining superseded information~\citep{chhikara2025mem0,xu2025amem,rasmussen2025zep}. Beyond storing reusable knowledge, recent systems also use memory to maintain an explicit view of ongoing task progress. MAGE maintains a hierarchical state tree with summary validation and rollback~\citep{chen2026mage}, while Ledger uses change counters to track observation freshness and reduce redundant actions~\citep{wang2026ledger}. These mechanisms help preserve execution state across long trajectories. In coding, however, maintaining task progress alone is insufficient: stored observations and test results may become stale as the repository changes. Recovering offloaded evidence therefore requires not only reconstructing what happened later, but also checking whether the evidence still applies to the current repository. MemTrace addresses this by binding immutable execution traces to repository state, recording verification, correction, and supersession relations in its Memory Trace Graph (MTG), and validating evidence before restoration.

\paragraph{Repository-level coding and code graphs.}
Repository-level coding systems coordinate localization, editing, and validation through agent--environment interaction or structured repair pipelines~\citep{yang2024sweagent,zhang2024autocoderover,xia2024agentless}. CodePlan further uses dependency analysis to plan coordinated changes across files~\citep{bairi2024codeplan}. CodexGraph, RepoGraph, and LocAgent represent code entities and their relations to support repository navigation and localization~\citep{liu2024codexgraph,ouyang2025repograph,chen2025locagent}. These representations expose structural dependencies that are difficult to recover from isolated code snippets. However, code structure alone does not record which observations supported an edit, which tests verified it, or whether later execution replaced an earlier conclusion. MemTrace links its Repository State Graph (RSG) to the MTG through shared repository entities, allowing relevant execution evidence to be located and checked against the evolving codebase. Table~\ref{tab:comparison} summarizes these differences across history retention, execution and repository structure, and evidence validity.

\begin{figure}[H]
    \vspace{-0.7em}
    \centering
    \includegraphics[
        width=\textwidth,
        trim=20pt 20pt 20pt 20pt,
        clip
    ]{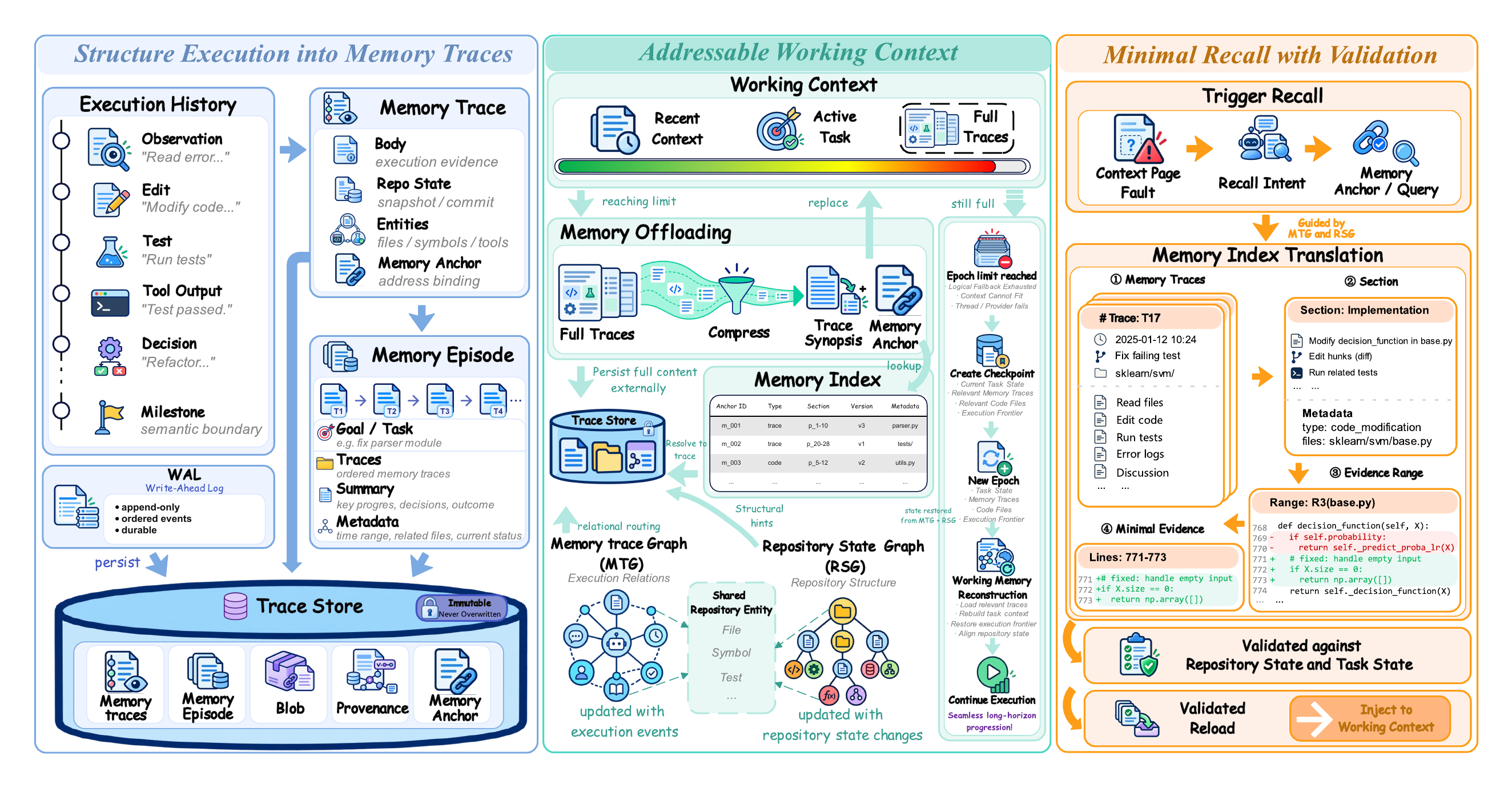}
    \caption{
    \textbf{Overview of MemTrace.}
    From left to right, MemTrace structures execution into state-bound traces,
    maintains an addressable working context, and restores only validated
    evidence needed to resume execution.
    }
    \label{fig:method-overview}
\end{figure}

\vspace{-0.7em}
\section{Method}
\label{sec:method}

MemTrace equips coding agents with persistent execution memory that survives context refreshes and supports evidence reuse as the repository evolves. It stores completed units of work as immutable \emph{Memory Traces} and records their execution order and dependencies in a \emph{Memory Trace Graph} (MTG). A complementary \emph{Repository State Graph} (RSG) represents the current files, symbols, tests, and their structural relations. Linking traces to repository entities allows MemTrace to check whether historical evidence remains applicable. When working memory approaches its context limit, MemTrace offloads completed trace bodies, retaining compact \emph{Memory Anchors} and lightweight \emph{Trace Synopses} that reference the persistent store. After a context refresh, it reconstructs the execution frontier, validates historical evidence against the current repository, and restores only what the next action requires.

\noindent\textbf{Workflow.} Figure~\ref{fig:method-overview} demonstrates the three stages of MemTrace. At step \(t\), MemTrace maintains the state
\(S_t=(R_t,\,m_t,\,\mathcal{H}_t,\,G^{\mathrm{M}}_t,\,G^{\mathrm{R}}_t,\,\Pi_t,\,C_t,\,\mathcal{A}_t)\).
Here \(R_t\) is the current repository state, including working-tree edits,
\(m_t\) records the execution frontier and unresolved work,
and \(\mathcal{H}_t\) is the set of immutable traces in the persistent
Trace Store.
\(G^{\mathrm{M}}_t\) and \(G^{\mathrm{R}}_t\) are the MTG and RSG;
\(\Pi_t\) associates traces with repository entities;
\(C_t\) is working memory; and \(\mathcal{A}_t\) contains the anchors
currently visible in it.
The usable input-context budget is \(B\), with space reserved for the
next response.
Instructions, task text, tool descriptions, checkpoint metadata, and
restored evidence all count toward this budget. We next present state-bound Memory Traces, the MTG–RSG alignment through shared repository entities, and validated restoration after a context refresh.


\subsection{State-Bound Memory Traces}
\label{sec:traces}

Long execution histories consume context, while repository changes can invalidate earlier observations. To support compact retention and reliable reuse, MemTrace organizes history into immutable \emph{Memory Traces}, each binding a completed unit of execution evidence to the repository state in which it was produced. Trace construction and offloading proceed as follows.

\paragraph{Identify and seal reusable evidence.}
MemTrace seals a unit of work when it is complete and provides evidence that can support subsequent actions, such as an edit, test outcome, failure diagnosis, or correction. Each trace is represented as
\[
\tau =
\bigl(x_{\tau},\, \rho_{\tau},\, \mathcal{E}(\tau),\, \alpha_{\tau}\bigr),
\]
where \(x_{\tau}\) contains the execution evidence and its provenance;
\(\rho_{\tau}\) records the repository state at evidence creation;
\(\mathcal{E}(\tau)\) identifies the files, symbols, and tests referenced in the body;
and \(\alpha_{\tau}\) is a compact \emph{Memory Anchor} that provides a stable reference to the trace.
Sealing adds \(\tau\) to the persistent Trace Store \(\mathcal{H}_t\) and its anchor to \(\mathcal{A}_t\).
A unit that exceeds the size of a single trace body is split into multiple traces, each with its own anchor.

\paragraph{Preserve provenance as the repository evolves.}
Once sealed, a trace is never modified. Subsequent edits or corrections produce new traces and relations, preserving the original evidence, repository state, and entity bindings. These records make the conditions under which evidence was produced available for later validation. Persistence alone, however, does not establish current validity: a recorded test pass may no longer apply after the tested code changes. MemTrace therefore retains historical traces for provenance but reuses their evidence only when it remains applicable. Section~\ref{sec:restore} describes how \(\rho_{\tau}\) and \(\mathcal{E}(\tau)\) are checked against the current repository state \(R_t\) and task state \(m_t\) before restoration.

\paragraph{Offload bodies while retaining access.}
When working memory \(C_t\) approaches the context budget \(B\), \emph{Memory Offloading} removes selected completed trace bodies from the active context while preserving their full records in \(\mathcal{H}_t\). Compact \emph{Trace Synopses} summarize the offloaded content, and Memory Anchors provide stable references to it. Both count toward \(B\), so only a budgeted subset remains in \(C_t\). Anchors omitted from the active context remain stored with their traces and are recoverable through the persistent index. This separates persistent retention from in-context use: evidence remains accessible, while its restoration depends on the validation procedure in Section~\ref{sec:restore}.

\begin{figure}[t]
    \centering
    \includegraphics[width=0.75\linewidth]
    {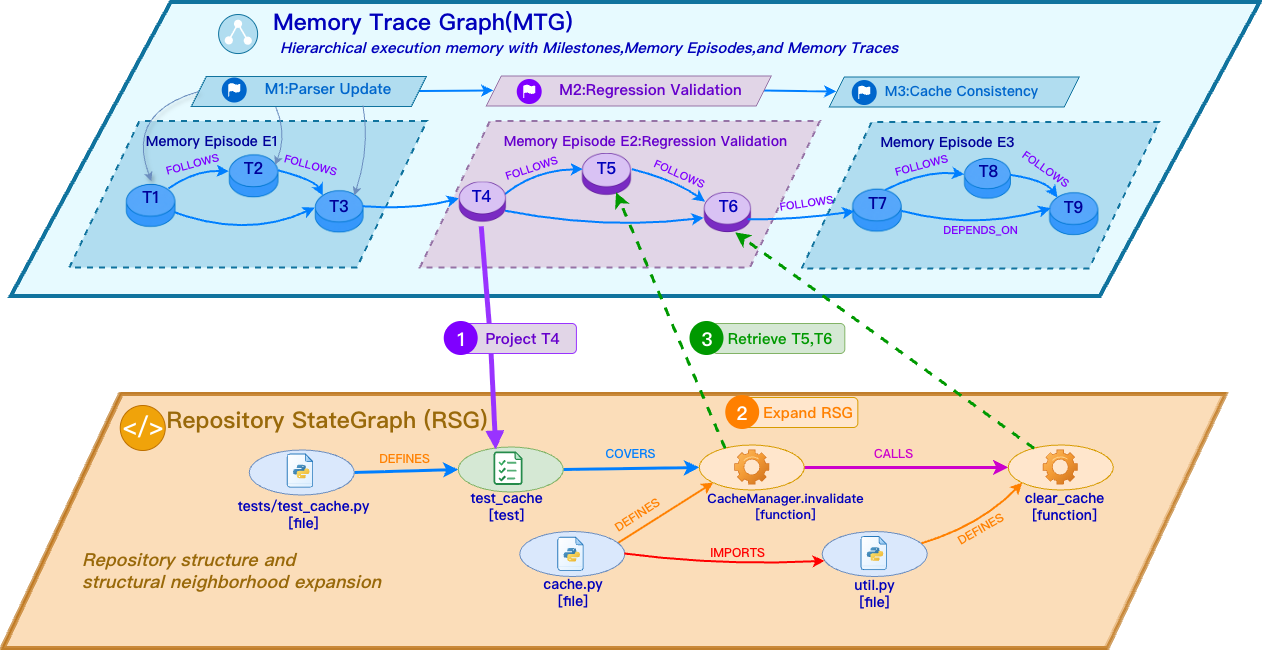}
    \caption{
    \textbf{MTG--RSG Projection and candidate localization.}
    Repository bindings connect execution traces to current repository
    entities, allowing MTG--RSG traversal to locate relevant traces for
    validated restoration.
    }
    \label{fig:mtg-rsg-alignment}
\end{figure}

\subsection{MTG--RSG Projection}
\label{sec:alignment}

MemTrace links execution history to the evolving codebase through two complementary graphs: the \emph{Memory Trace Graph} (MTG) records task progress, and the \emph{Repository State Graph} (RSG) represents current repository structure. Their nodes and edges remain separate; the projection \(\Pi_t\) connects traces to repository entities.

\paragraph{Execution and repository graphs.}
The MTG \(G^{\mathrm{M}}_t\) contains one node per trace in \(\mathcal{H}_t\), with relations recording execution order, dependencies, verification, correction, and supersession. For example, a repair trace may depend on a failure diagnosis and supersede an earlier edit, while a passing-test trace verifies the repair. These relations identify evidence that supports or revises the frontier \(m_t\).

The RSG \(G^{\mathrm{R}}_t\) connects task-relevant files, symbols, and tests through \emph{defines}, \emph{calls}, \emph{imports}, and \emph{covers} relations. Here, \emph{covers} denotes a test-to-code association for localization, not measured coverage or a passing result. MemTrace incrementally updates the graph around the execution frontier as \(R_t\) changes. Its structural relations guide retrieval without establishing the validity of historical evidence.

\paragraph{Trace--repository projection.}
Each entity reference identifies a file, symbol, or test within its repository scope; symbols include their file path and qualified name. Sealing a trace records its bindings in
\begin{equation}
\Pi_t =
\bigl\{(\tau,e)\mid
\tau\in\mathcal{H}_t,\;
e\in\mathcal{E}(\tau)
\bigr\}.
\label{eq:projection}
\end{equation}
An anchor resolves to one trace, whereas an entity lookup may return several. In Figure~\ref{fig:mtg-rsg-alignment}, a cache-invalidation query starts from \(T_4\), which references \texttt{test\_cache}. RSG traversal follows \emph{covers} to \texttt{CacheManager.invalidate} and \emph{calls} to \texttt{clear\_cache}; lookups through \(\Pi_t\) then retrieve the associated traces \(T_5\) and \(T_6\).

\paragraph{Candidate retrieval.}
Retrieval starts from anchored traces, traces supporting \(m_t\), or repository references. RSG expansion follows a bounded, request-specific sequence of relations; the figure's two-relation path is illustrative, not a global limit. Candidates may then expand by at most two MTG hops along dependency, verification, correction, and supersession relations. Traversal inspects incoming and outgoing edges without changing their meaning, allowing a failure trace to locate a later repair. Direct candidates precede expanded neighbors, and duplicates are removed before validation.

\subsection{Validated Restoration}
\label{sec:restore}

MemTrace restores evidence only when it is relevant to the next action, applicable to the current execution state, and within the context budget.

\paragraph{Applicability check.}
A trace is applicable when
\begin{equation}
\mathrm{applies}(\tau;R_t,m_t)
\iff
\mathrm{aligned}(\tau,R_t)
\;\wedge\;
\neg\mathrm{superseded}(\tau,m_t).
\label{eq:applies}
\end{equation}
The predicate \(\mathrm{aligned}\) checks the evidence's recorded source conditions against \(R_t\), using \(\rho_{\tau}\), \(\mathcal{E}(\tau)\), and provenance in \(x_{\tau}\). It does not require the entire repository to remain unchanged. For a test result, however, both the test and its recorded dependencies must be checked; an unchanged target function alone is insufficient. Changed or unverifiable conditions prevent automatic reuse. The predicate \(\mathrm{superseded}\) requires an explicit MTG relation indicating replacement for the current frontier; temporal order alone is insufficient.

\paragraph{Selective restoration.}
MemTrace inspects retrieved candidates and, if needed, expands them as described in Section~\ref{sec:alignment}. Relevant traces satisfying Eq.~\eqref{eq:applies} are restored with their source labels and anchors until the request is satisfied. Blocks that exceed the remaining budget are deferred rather than truncated. Explicit historical queries may retrieve older or superseded traces labeled as historical evidence. If applicable evidence is unavailable, the agent reads the current file or symbol, or reruns the required test. Budget-deferred evidence is tracked separately from unavailable or invalid evidence.

\paragraph{Recovery after a context refresh.}
Before refreshing the context, MemTrace offloads completed bodies and checkpoints \(m_t\) and the anchors for active work; the Trace Store, both graphs, and \(\Pi_t\) persist. Recovery then proceeds in three steps:
(1) reconstruct task progress from MTG traces supporting or revising the frontier;
(2) use \(\Pi_t\) and the RSG to locate repository entities relevant to continuation; and
(3) restore relevant candidates that pass validation and fit within \(B\).
The admitted evidence, original task, recovered frontier, and current repository scope form the refreshed working memory. Algorithm~\ref{alg:context-refresh} summarizes this procedure.

\begin{algorithm}[t]
\caption{MTG--RSG context recovery}
\label{alg:context-refresh}
\begin{algorithmic}[1]
\Require State \(S_t\), recovery request \(q\), context budget \(B\)
\Ensure Reconstructed memory \(C_t^{+}\)

\State Checkpoint \(m_t\) and \(\mathcal{A}_t\); offload completed bodies
\State \(Q \gets\) anchored or frontier-related traces
\State Expand \(Q\) by up to two MTG hops
\State Add \(q\)-relevant traces via the RSG and \(\Pi_t\)
\State Initialize \(C_t^{+}\) with the task, checkpoint, and repository scope

\For{each \(q\)-relevant trace \(\tau \in Q\)}
    \If{\(\mathrm{applies}(\tau;R_t,m_t)\) and
        its body and anchor fit the remaining budget}
        \State Restore \(x_{\tau}\) and \(\alpha_{\tau}\) into \(C_t^{+}\)
    \EndIf
\EndFor
\State \Return \(C_t^{+}\)
\end{algorithmic}
\end{algorithm}
\section{Experiments}
\label{sec:exp}

We evaluate \textsc{MemTrace} on three long-horizon coding benchmarks and two agent harnesses.
We compare task performance (Section~\ref{sec:main_results}) and analyze component contributions and memory behavior (Section~\ref{sec:analysis}). Wall-clock results are in Appendix~\ref{app:exp-details}.

\subsection{Experimental Setup}
\label{sec:exp_setup}

\paragraph{Benchmarks.}
We evaluate on three long-horizon software engineering benchmarks covering complementary task settings.
DeepSWE~\citep{huang2026deepswe} contains 113 tasks across 91 repositories and five programming languages.
SWE-EVO~\citep{thai2025sweevo} comprises 48 software-evolution tasks constructed from release histories of seven Python projects. Two instances already pass all FAIL\_TO\_PASS tests without code changes, so we evaluate the remaining 46 tasks.
SWE-Milestone~\citep{swemilestone2026catalogue} contains 98 human-verified milestones across seven continuous software-evolution itineraries spanning five languages.

\paragraph{Backbone, harnesses, and baselines.}
We use \texttt{deepseek-v4-flash-}\allowbreak\texttt{0731-high}
for all experiments, with a 200K-token context window and no explicit step limit.
Experiments are conducted under Codex CLI v0.144.4~\citep{bolin2026codex}
and mini-swe-agent v2.4.6~\citep{minisweagent2025}. We compare against three baselines.
Native uses the default context management of each harness. Observation Masking~\citep{lindenbauer2025complexitytrap}
masks older tool observations outside a rolling window while retaining the reasoning and action history. RepoGraph~\citep{ouyang2025repograph}
augments the agent with repository-level structural retrieval and is evaluated only on Python tasks, following its implementation scope.

\paragraph{Metrics.}
We report benchmark-specific effectiveness metrics together with wall-clock time.
For DeepSWE, we report pass@1 and the average wall-clock time per task, where pass@1 measures the fraction of tasks successfully resolved.
For SWE-EVO, we report Resolved Rate, Fix Rate, and Apply, measuring complete resolution, partial functional progress, and successful patch application, respectively~\citep{thai2025sweevo}; we additionally report the average wall-clock time per task.
For SWE-Milestone, we report Score, Precision, Recall, and Resolved Rate. Score is the harmonic mean of Precision and Recall, which measure functional correctness with respect to regressions and the fraction of required functionality completed, respectively~\citep{swemilestone2026catalogue}.

\subsection{Main Results}
\label{sec:main_results}

Table~\ref{tab:main} compares \textsc{MemTrace} with the baselines across
three benchmarks and two agent harnesses.
Beyond the overall gains, the results reveal three notable patterns:
strong harness sensitivity among the baselines, persistent gains in
Python-only settings, and a clear distinction between partial progress
and complete task resolution.

\begin{table}[t]
\caption{Main results (\%) on DeepSWE, SWE-EVO, and SWE-Milestone.
Full-benchmark results are reported with Python subsets in parentheses;
SWE-EVO is Python-only.
\textbf{Bold} denotes \textsc{MemTrace}, and \underline{underlining}
denotes the best baseline.
$\Delta_{\mathrm{best}}$ reports the percentage-point difference from the best baseline,
computed directly from the reported values.}
\label{tab:main}
\centering
\small
\setlength{\tabcolsep}{3.2pt}
\renewcommand{\arraystretch}{1.12}
\resizebox{\linewidth}{!}{
\begin{tabular}{@{}lcccccccc@{}}
\toprule
& \multicolumn{1}{c}{\textbf{DeepSWE}}
& \multicolumn{3}{c}{\textbf{SWE-EVO}}
& \multicolumn{4}{c}{\textbf{SWE-Milestone}} \\
\cmidrule(lr){2-2}
\cmidrule(lr){3-5}
\cmidrule(l){6-9}
\textbf{Method}
& pass@1 & Resolved & Fix & Apply
& Score & Prec. & Recall & Resolved \\
\midrule

\rowcolor{MTBlue}
\multicolumn{9}{c}{\textbf{Harness: Codex CLI}} \\

Native
& \mtbest{35.4} (32.4)
& 47.8
& 53.4
& \mtbest{97.8}
& \mtbest{28.6} (30.7)
& \mtbest{26.9} (26.4)
& \mtbest{41.9} (84.5)
& 7.1 (\mtbest{8.3}) \\

Obs.\ Masking
& 34.5 (38.2)
& \mtbest{65.2}
& \mtbest{72.1}
& \mtbest{97.8}
& 24.4 (30.0)
& 24.2 (26.0)
& 37.1 (84.9)
& \mtbest{9.2} (\mtbest{8.3}) \\

RepoGraph
& -- (\mtbest{41.2})
& 58.7
& 68.8
& \mtbest{97.8}
& -- (\mtbest{45.2})
& -- (\mtbest{36.7})
& -- (\mtbest{91.5})
& -- (\mtbest{8.3}) \\

\textbf{\textsc{MemTrace}}
& \textbf{56.6 (67.6)}
& \textbf{69.6}
& \textbf{70.6}
& \textbf{100.0}
& \textbf{46.4 (46.1)}
& \textbf{43.6 (37.8)}
& \textbf{60.9 (91.7)}
& \textbf{16.3 (16.7)} \\

\cmidrule(lr){1-9}
$\Delta_{\mathrm{best}}$
& \mtgain{+21.2 (+26.4)}
& \mtgain{+4.4}
& \mtloss{-1.5}
& \mtgain{+2.2}
& \mtgain{+17.8 (+0.9)}
& \mtgain{+16.7 (+1.1)}
& \mtgain{+19.0 (+0.2)}
& \mtgain{+7.1 (+8.4)} \\

\addlinespace[2pt]

\rowcolor{MTBlue}
\multicolumn{9}{c}{\textbf{Harness: mini-swe-agent}} \\

Native
& \mtbest{48.7} (\mtbest{47.1})
& \mtbest{52.2}
& \mtbest{54.1}
& 95.7
& \mtbest{34.5} (17.6)
& \mtbest{34.7} (19.6)
& \mtbest{45.6} (54.2)
& \mtbest{11.2} (0.0) \\

Obs.\ Masking
& 26.6 (20.6)
& 23.9
& 33.6
& \mtbest{97.8}
& 24.9 (21.9)
& 25.1 (21.4)
& 29.9 (22.5)
& 8.2 (\mtbest{8.3}) \\

RepoGraph
& -- (32.4)
& 30.4
& 37.9
& \mtbest{97.8}
& -- (\mtbest{30.8})
& -- (\mtbest{25.5})
& -- (\mtbest{69.1})
& -- (0.0) \\

\textbf{\textsc{MemTrace}}
& \textbf{66.4 (64.7)}
& \textbf{63.0}
& \textbf{68.7}
& \textbf{93.5}
& \textbf{48.3 (37.3)}
& \textbf{46.9 (32.6)}
& \textbf{71.2 (82.2)}
& \textbf{19.4 (16.7)} \\

\cmidrule(lr){1-9}
$\Delta_{\mathrm{best}}$
& \mtgain{+17.7 (+17.6)}
& \mtgain{+10.8}
& \mtgain{+14.6}
& \mtloss{-4.3}
& \mtgain{+13.8 (+6.5)}
& \mtgain{+12.2 (+7.1)}
& \mtgain{+25.6 (+13.1)}
& \mtgain{+8.2 (+8.4)} \\

\bottomrule
\end{tabular}
}
\end{table}

\textbf{(i) Baseline effectiveness depends strongly on the harness, whereas
\textsc{MemTrace} remains effective across both.}
\textsc{MemTrace} achieves the best primary metric in every
benchmark--harness setting.
In contrast, the relative performance of the baselines changes substantially
across harnesses. On SWE-EVO, for example, Observation Masking improves over
Native by 17.4 points under Codex CLI, but falls 28.3 points below Native
under mini-swe-agent; RepoGraph shows a similar reversal.
This suggests that context reduction and structural retrieval are sensitive
to the surrounding agent execution policy, while the benefit of
\textsc{MemTrace} is more stable across harnesses.

\textbf{(ii) The gain persists on Python, where RepoGraph is directly applicable.}
On the DeepSWE Python subset, \textsc{MemTrace} exceeds RepoGraph by more
than 26 points under Codex CLI and 32 points under mini-swe-agent.
A different pattern appears on SWE-Milestone: under Codex CLI,
\textsc{MemTrace} and RepoGraph are nearly tied on Python Score, Precision,
and Recall, yet the strict Resolved Rate increases from 8.3\% to 16.7\%.
Thus, the advantage is not explained by language coverage alone; on
sequential evolution tasks, it is particularly visible in converting
partial progress into complete solutions.

\textbf{(iii) Higher partial-progress metrics do not necessarily translate
into higher task resolution.}
On SWE-EVO under Codex CLI, Observation Masking attains a slightly higher
Fix Rate than \textsc{MemTrace} (72.1\% vs.\ 70.6\%), but a lower Resolved
Rate (65.2\% vs.\ 69.6\%).
Under mini-swe-agent, \textsc{MemTrace} has a lower Apply rate than the best
baseline (93.5\% vs.\ 97.8\%) while achieving a substantially higher
Resolved Rate (63.0\% vs.\ 52.2\%).
The improvement therefore appears primarily in end-to-end task completion,
rather than in patch applicability or partial repair alone.

\subsection{Analysis}
\label{sec:analysis}

\paragraph{Component ablation.}
Adding the MTG to the Trace Store improves DeepSWE from 32.1\% to 44.2\%, showing that execution relations help connect earlier edits, tests, and corrections across a long trajectory.
Adding the RSG further raises performance to 56.6\%, indicating the value of tracking relations among files, symbols, and tests when task progress spans a broad repository context.
This effect is much smaller on SWE-EVO, where removing the RSG leaves performance unchanged at 69.6\%.
One likely reason is the difference in task scope: DeepSWE covers repository-level tasks across many repositories and languages, whereas SWE-EVO follows release-level evolution within a smaller set of projects.
As a result, explicit repository grounding appears more useful when recovery must reconnect evidence across a broader and more heterogeneous codebase.

\begin{figure}[!t]
    \centering
    \includegraphics[width=.8\linewidth]
        {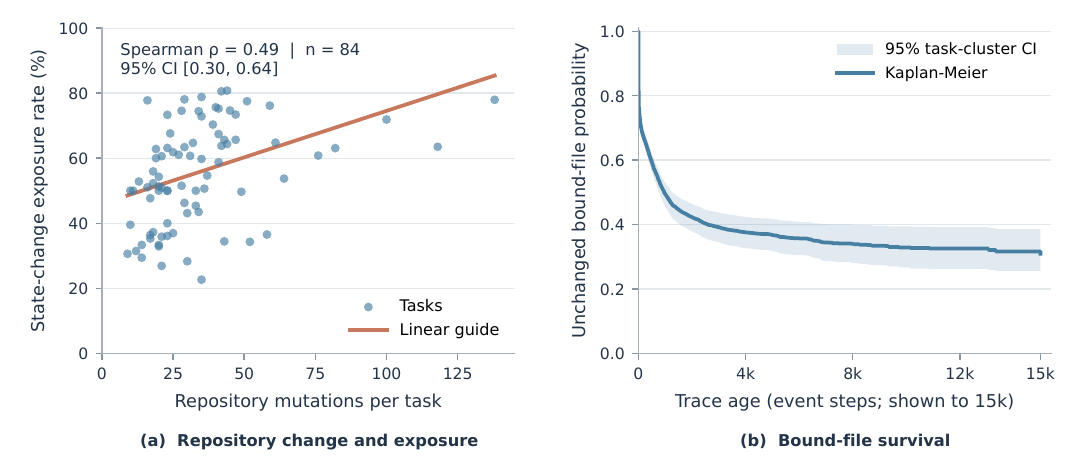}
    \caption{State--memory alignment under repository evolution.
    (a) Each point is one task, showing the relation between repository revisions and the fraction of Memory Traces whose bound files are later modified.
    (b) The probability that a trace's bound files remain unchanged as the trace becomes older.
    The shaded region shows the 95\% task-cluster bootstrap confidence interval.
    Details are in Appendix~\ref{app:state-analysis}.}
    \label{fig:state-memory-alignment}
\end{figure}

\begin{table}[t]
\caption{Component ablation under Codex CLI.
Trace Store Only replaces MTG and RSG routing with flat retrieval over Trace Synopses and bodies;
w/o MTG removes execution-order and dependency edges;
w/o RSG removes repository-structure candidate expansion.
Cells report Resolved Rate for SWE-EVO and pass@1 for DeepSWE.}
\label{tab:ablation}
\centering
\small
\setlength{\tabcolsep}{5pt}
\renewcommand{\arraystretch}{1.08}
\begin{tabular}{@{}lccccc@{}}
\toprule
\multirow{2}{*}{\textbf{Variant}}
& \multicolumn{3}{c}{\textbf{Component}}
& \multirow{2}{*}{\shortstack{\textbf{SWE-EVO}\\Resolved Rate (\%)}}
& \multirow{2}{*}{\shortstack{\textbf{DeepSWE}\\pass@1 (\%)}} \\
\cmidrule(lr){2-4}
& Trace Store
& MTG
& RSG
&
& \\
\midrule
Native
& --
& --
& --
& 47.8
& 35.4 \\
Trace Store Only
& \checkmark
& --
& --
& 65.2
& 32.1 \\
\textsc{MemTrace} w/o MTG
& \checkmark
& --
& \checkmark
& 65.2
& 34.5 \\
\textsc{MemTrace} w/o RSG
& \checkmark
& \checkmark
& --
& \textbf{69.6}
& 44.2 \\
\textsc{MemTrace}
& \checkmark
& \checkmark
& \checkmark
& \textbf{69.6}
& \textbf{56.6} \\
\bottomrule
\end{tabular}
\end{table}
\paragraph{State--memory alignment.}
We examine how later repository changes affect stored execution evidence.
Among 9,226 eligible file-bound Memory Traces, 57.0\% encounter a later change to at least one bound file.
Exposure is more frequent in tasks with more repository revisions
(Spearman $\rho=0.49$; Figure~\ref{fig:state-memory-alignment}(a)).
The estimated probability that bound files remain unchanged falls to
about 0.50 after 1,000 logged event steps
(Figure~\ref{fig:state-memory-alignment}(b)).
These observations motivate checking current applicability before reusing
historical evidence. Definitions and statistics are in
Appendix~\ref{app:state-analysis}.

\paragraph{Long-horizon behavior.}
We group DeepSWE tasks into short, medium, and long execution horizons using response counts from an independent Native run (Appendix~\ref{app:horizon-analysis}).
Across the three groups, median stored Memory Traces increase as \(79 \rightarrow 102.5 \rightarrow 129\), repository revisions as \(21 \rightarrow 30 \rightarrow 41\), and semantic evidence invalidations as \(122 \rightarrow 142 \rightarrow 223\) (Figure~\ref{fig:long-horizon-behavior}(a--c)).
Longer execution therefore corresponds to more accumulated history and repository evolution, increasing state-aware memory maintenance.
The exposure rate, however, shows no clear monotonic increase with execution length (Figure~\ref{fig:long-horizon-behavior}(d)), indicating that the added maintenance load is driven mainly by more stored history and repository changes rather than a higher per-trace exposure rate.

\FloatBarrier
\section{Conclusion and Future Work}
\label{sec:conclusion}

We introduced \textsc{MemTrace}, a state-consistent memory system for
long-horizon coding agents that preserves execution evidence with repository
provenance and validates it before reuse.
Across long-horizon benchmarks, MemTrace consistently improves
the primary task-level metric over fully evaluated baselines across two
agent harnesses, while our analyses show the complementary value of
structured execution memory and repository-state grounding.
These results suggest that long-horizon memory should maintain not only
historical information, but also its applicability to the current execution
state.
Future work will evaluate MemTrace with additional backbone models and
context-window sizes, and extend it to persistent development settings
spanning multiple tasks and sessions within the same repository.

\clearpage
\TracePageBibliography
\clearpage
\appendix
\section{Appendix}

\makeatletter
\setlength{\@fptop}{0pt}
\setlength{\@fpsep}{10pt plus 4pt minus 2pt}
\setlength{\@fpbot}{0pt plus 1fil}
\makeatother
\setlength{\floatsep}{8pt plus 2pt minus 2pt}
\setlength{\textfloatsep}{10pt plus 2pt minus 4pt}

\Needspace{5\baselineskip}
\subsection{Benchmark Composition}
\label{app:benchmarks}

Table~\ref{tab:benchmark-languages} summarizes the official benchmark composition.
DeepSWE contains 113 tasks from 91 repositories~\citep{huang2026deepswe};
SWE-EVO contains 48 tasks from seven Python projects~\citep{thai2025sweevo}.
SWE-Milestone comprises seven repository-level itineraries and
98 graded milestones~\citep{swemilestone2026catalogue}.
Milestones are graded stages within an itinerary.

\begin{table}[H]
\centering
\small
\setlength{\tabcolsep}{7pt}
\renewcommand{\arraystretch}{1.08}
\caption{Official task distribution by repository language~\citep{huang2026deepswe,thai2025sweevo,swemilestone2026catalogue}. SWE-Milestone reports both evolution itineraries and graded milestones.}
\label{tab:benchmark-languages}
\begin{tabular}{@{}lrrrr@{}}
\toprule
& \textbf{DeepSWE} & \textbf{SWE-EVO} & \multicolumn{2}{c}{\textbf{SWE-Milestone}} \\
\cmidrule(lr){4-5}
\textbf{Language} & Tasks & Tasks & Itineraries & Graded milestones \\
\midrule
Python     & 34 & 48 & 1 & 12 \\
TypeScript & 35 & 0  & 1 & 18 \\
Go         & 34 & 0  & 2 & 32 \\
JavaScript & 5  & 0  & 0 & 0  \\
Rust       & 5  & 0  & 2 & 24 \\
Java       & 0  & 0  & 1 & 12 \\
\midrule
\textbf{Total} & \textbf{113} & \textbf{48} & \textbf{7} & \textbf{98} \\
\bottomrule
\end{tabular}
\end{table}

\paragraph{Counting convention.}
Graded milestone counts are go-zero (23), Navidrome (9), element-web (18),
Nushell (13), ripgrep (11), Dubbo (12), and scikit-learn (12).
Language groups follow the official catalogue, which classifies
Navidrome under Go~\citep{swemilestone2026catalogue}.
Resolved-rate tables below use these official sets, with two stated exclusions.
SWE-EVO rates use 46 tasks: \code{pydantic} v2.7.1 to v2.7.2 and
\code{scikit-learn} 0.21.1 to 0.21.2 fail benchmark validation and are
removed before scoring.
They are not unresolved agent attempts.
Every DeepSWE rate in Table~\ref{tab:counts} uses all 113 tasks.

\Needspace{5\baselineskip}
\subsection{Additional Results and Wall-Clock Time}
\label{app:exp-details}

Table~\ref{tab:counts} reports the resolved-task counts underlying the main results.

\begin{table}[H]
\centering
\small
\setlength{\tabcolsep}{4pt}
\renewcommand{\arraystretch}{1.05}
\caption{Resolved-task counts for Table~\ref{tab:main}.}
\label{tab:counts}
\begin{tabular}{@{}llcccc@{}}
\toprule
& & \multicolumn{3}{c}{\textbf{DeepSWE}}
& \textbf{SWE-EVO} \\
\cmidrule(lr){3-5}
\cmidrule(l){6-6}
\textbf{Harness}
& \textbf{Method}
& All (113)
& Python (34)
& Non-Python (79)
& All (46) \\
\midrule
Codex CLI
& Native
& 40 & 11 & 29 & 22 \\
& Obs.\ Masking
& 39 & 13 & 26 & 30 \\
& RepoGraph
& -- & 14 & -- & 27 \\
& \textsc{MemTrace}
& 64 & 23 & 41 & 32 \\
\midrule
mini-swe-agent
& Native
& 55 & 16 & 39 & 24 \\
& Obs.\ Masking
& 30 & 7 & 23 & 11 \\
& RepoGraph
& -- & 11 & -- & 14 \\
& \textsc{MemTrace}
& 75 & 22 & 53 & 29 \\
\bottomrule
\end{tabular}
\end{table}

\Needspace{8\baselineskip}
RepoGraph is reported for Python tasks only.
SWE-Milestone uses its official itinerary-level aggregation.
The DeepSWE Trace Store Only ablation was scheduled on the same 113 tasks.
\code{httpx-streaming-json-iteration} has no official score
(\code{MISSING_RESULT}; the attempt exited in about 33 seconds) and is
excluded from both the numerator and the denominator.
The scored result is 36/112 (32.1\%).
That task is not counted as a resolved success or as a model failure.

\paragraph{Implementation scope.}
\label{app:limits}
Configurations share the backbone within each harness setup.
Component ablations use Codex CLI.
The audited RSG provides symbol-level relations for Python and
file-level relations for the other languages.

\paragraph{Wall-clock time.}
Table~\ref{tab:wallclock} reports mean elapsed time per task in minutes,
rounded to two decimal places.
Times are compared within each harness and benchmark.

\newsavebox{\wallclockbox}
\begin{table}[H]
\centering
\small
\setlength{\tabcolsep}{5pt}
\renewcommand{\arraystretch}{1.05}
\caption{Mean wall-clock time in minutes per task.}
\label{tab:wallclock}
\sbox{\wallclockbox}{%
\begin{tabular}{@{}lrrrr@{}}
\toprule
& \multicolumn{2}{c}{\textbf{DeepSWE}} & \multicolumn{2}{c}{\textbf{SWE-EVO}} \\
\cmidrule(lr){2-3}\cmidrule(l){4-5}
\textbf{Method} & Codex CLI & mini-swe-agent & Codex CLI & mini-swe-agent \\
\midrule
Native & 42.02 & 103.77 & 30.00 & 170.97 \\
Obs.\ Masking & 44.32 & 120.98 & 52.72 & 170.02 \\
RepoGraph & 34.67$^{\dagger}$ & 70.43$^{\dagger}$ & 34.22 & 203.28 \\
\textsc{MemTrace} & 54.17 & 75.30 & 82.05 & 114.03 \\
\bottomrule
\end{tabular}}%
\usebox{\wallclockbox}\par
\vspace{3pt}
\begin{minipage}{\wd\wallclockbox}
\centering
\footnotesize
\textit{Note.} $^{\dagger}$ DeepSWE RepoGraph means use the 34 Python tasks. Other DeepSWE means use 113 tasks.
\end{minipage}
\end{table}

\Needspace{5\baselineskip}
\subsection{State--Memory Alignment}
\label{app:state-analysis}

\paragraph{Analysis population.}
These diagnostics use archived DeepSWE mini-swe-agent trajectories,
separately from the performance runs in Table~\ref{tab:main}.
Exposure analysis includes 9,226 Memory Traces from 84 trajectories,
each with a recorded sealing event and nonempty file bindings.
Revision, invalidation, and Context Refresh counts use 85 instrumented
trajectories (Table~\ref{tab:alignment-coverage}).
Memory Trace identifiers and semantic evidence records are counted separately.

\begin{table}[H]
\centering
\small
\setlength{\tabcolsep}{5pt}
\caption{Analysis population and event counts for the state--memory diagnostics.}
\label{tab:alignment-coverage}
\begin{tabular}{@{}lr@{}}
\toprule
\textbf{Quantity} & \textbf{Count} \\
\midrule
Event-instrumented trajectories & 85 \\
Trajectories with eligible file-bound Memory Traces & 84 \\
Eligible time-aligned, file-bound Memory Traces & 9,226 \\
Memory Traces with a later bound-file change & 5,261 \\
Repository revision events & 2,897 \\
Semantic evidence invalidations & 16,197 \\
Explicit context-refresh events & 258 \\
\bottomrule
\end{tabular}
\end{table}

\Needspace{6\baselineskip}
\paragraph{Event clock and repository revisions.}
For task $q$, let $u_q$ be its last recorded event step and
$\mathcal{V}_q$ its revision events.
Each revision $r$ has a step $s_r$ and modified-file set $\Delta F_r$.
The sealing step $c_\tau$ is the recorded finalization of Memory Trace $\tau$,
matched by its identifier.
Repository revisions count \codepath{WORKSPACE_REVISION_ADVANCED} events:
\begin{equation}
    M_q = |\mathcal{V}_q|.
    \label{eq:diag-revisions}
\end{equation}
The figure label ``repository mutations'' refers to the same count.
Event steps are log positions, distinct from model responses and elapsed time.

\paragraph{Repository-state exposure.}
Let $F_\tau$ be the files bound to Memory Trace $\tau$ at sealing,
and $\mathcal{T}^{*}_q$ the traces with a recorded sealing step and
nonempty $F_\tau$.
Exposure records a later change to at least one bound file:
\begin{equation}
    x_\tau = \mathbf{1}\!\left[
        \exists r\in\mathcal{V}_q:
        c_\tau < s_r \leq u_q
        \ \land\ F_\tau\cap\Delta F_r\neq\varnothing
    \right].
    \label{eq:diag-exposure-indicator}
\end{equation}
The task-level and pooled rates are
\begin{equation}
    E_q =
    \frac{
        \sum_{\tau\in\mathcal{T}^{*}_q}x_\tau
    }{
        |\mathcal{T}^{*}_q|
    },
    \qquad
    E =
    \frac{
        \sum_q\sum_{\tau\in\mathcal{T}^{*}_q}x_\tau
    }{
        \sum_q|\mathcal{T}^{*}_q|
    }.
    \label{eq:diag-exposure-rate}
\end{equation}
The pooled rate weights traces equally; Figure~\ref{fig:state-memory-alignment}(a)
uses one $(M_q,E_q)$ pair per eligible task.
The pooled rate is $5{,}261/9{,}226=57.0\%$
(95\% CI: 52.8--61.3\%).
Exposure measures file changes; Trace Validation assesses whether
historical evidence remains applicable.

\Needspace{7\baselineskip}
\paragraph{Semantic evidence invalidations.}
\codepath{SemanticStore.advance_workspace_revision} in
\codepath{semantic_memory/store.py} counts evidence identifiers selected
for closure at each revision.
Let $\mathcal{C}_{r^-}$ contain records marked \code{CURRENT} with an open
validity interval (\code{valid_to_cursor IS NULL}) before revision $r$.
Write $\nu(e)$ for a record's revision, $\nu_r$ for the new revision,
and $A_r(e)$ for the affected-entity predicate.
For global fact types $\mathcal{G}$ and entity-scoped types $\mathcal{D}$,
the selected identifiers are
\begin{equation}
\begin{split}
    \mathcal{I}_r = \{\operatorname{id}(e):\;&
        e\in\mathcal{C}_{r^-},\ \nu(e)\neq\nu_r,\\
    &\operatorname{type}(e)\in\mathcal{G}\ \lor\
    [\operatorname{type}(e)\in\mathcal{D}\land A_r(e)]\}.
\end{split}
\label{eq:diag-invalidated-set}
\end{equation}
Per-revision and cumulative per-task counts are
\begin{equation}
    I_r = |\mathcal{I}_r|,
    \qquad
    I_q = \sum_{r\in\mathcal{V}_q} I_r.
    \label{eq:diag-invalidation-count}
\end{equation}
Global types are \codepath{TOOL_RESULT}, \codepath{TEST_RESULT},
\codepath{TEST_FAILURE}, \codepath{VERIFIER_RESULT}, and
\codepath{REQUIREMENT_REVIEW}; they close on a revision regardless of
modified-file overlap.
Entity-scoped types, \codepath{CODE_CHANGE} and \codepath{CODE_OBSERVATION},
also require \codepath{_entity_affected} to match a file, symbol, or test binding.
Closure records the validity endpoint and an \code{INVALIDATED_BY}
relation while retaining the Memory Trace.
The 2,897 revisions yield 16,197 recorded invalidations.
Thus, $I_q$ measures revision-triggered maintenance under this policy,
not independently verified semantic errors.

\paragraph{Statistical procedure.}
We report task-level Spearman correlations and 95\% percentile bootstrap
intervals from 5,000 resamples.
Trace-level rates and survival resample tasks with all their eligible traces;
correlations resample task pairs; Long--Short median differences resample
within each group.
The scripts use base seed 20260926 with fixed per-analysis offsets.
Table~\ref{tab:alignment-associations} reports the task-level associations.

\begin{table}[!htbp]
\centering
\small
\setlength{\tabcolsep}{5pt}
\caption{Task-level associations with 95\% task-bootstrap confidence intervals. Pearson correlation is a secondary descriptive statistic.}
\label{tab:alignment-associations}
\begin{tabular}{@{}lccc@{}}
\toprule
\textbf{Variables}
& $N$
& \textbf{Spearman $\rho$ [95\% CI]}
& \textbf{Pearson $r$} \\
\midrule
Revisions vs. exposure
& 84
& $0.487\ [0.305,0.642]$
& 0.394 \\
Revisions vs. invalidations
& 85
& $0.837\ [0.741,0.899]$
& 0.881 \\
\bottomrule
\end{tabular}
\end{table}

\paragraph{Time to the first bound-file change.}
For an eligible Memory Trace, define
\begin{equation}
    T_\tau =
    \min\left\{
        s_r-c_\tau:
        r\in\mathcal{V}_q,\ s_r>c_\tau,\
        F_\tau\cap\Delta F_r\neq\varnothing
    \right\},
    \label{eq:diag-first-change}
\end{equation}
with $T_\tau=\infty$ when no change is observed.
Follow-up ends at $C_\tau=u_q-c_\tau$.
We observe $Y_\tau=\min(T_\tau,C_\tau)$ and
$\delta_\tau=\mathbf{1}[T_\tau\leq C_\tau]$;
traces without a recorded change are right-censored at the run endpoint.
At event age $a_j$, let $d_j$ count first changes and $n_j$ count traces
still observed and unchanged immediately beforehand.
The Kaplan--Meier estimate is
\begin{equation}
    \widehat S(a)=
    \prod_{a_j\leq a}
    \left(1-\frac{d_j}{n_j}\right).
    \label{eq:diag-km}
\end{equation}
Figure~\ref{fig:state-memory-alignment}(b) estimates the probability that
no bound file has yet changed, using the internal event clock and pointwise
task-cluster bootstrap intervals.
Table~\ref{tab:bound-state-survival} gives selected values.

\begin{table}[!htbp]
\centering
\small
\setlength{\tabcolsep}{6pt}
\caption{Bound-file survival at selected event-step ages. Intervals are 95\% task-cluster bootstrap intervals; at-risk counts are measured immediately before each age.}
\label{tab:bound-state-survival}
\begin{tabular}{@{}rcr@{}}
\toprule
\textbf{Event-step age}
& $\widehat S(a)$ \textbf{[95\% CI]}
& \textbf{Memory Traces at risk} \\
\midrule
97
& $0.703\ [0.682,0.723]$
& 6,380 \\
1,001
& $0.495\ [0.456,0.530]$
& 3,654 \\
2,012
& $0.422\ [0.379,0.464]$
& 2,404 \\
\bottomrule
\end{tabular}
\end{table}

Grouping traces by total observed age $C_\tau$ gives exposure rates of
27.4\%, 60.8\%, 69.9\%, and 70.1\% across quartiles.
These are descriptive summaries with different follow-up lengths;
the time-to-change estimate accounts for right censoring.

\Needspace{5\baselineskip}
\subsection{Long-Horizon Behavior}
\label{app:horizon-analysis}

\paragraph{Execution-length groups.}
For task $q$, execution length is the number of distinct model responses
in an independent Native mini-swe-agent trajectory:
\begin{equation}
    H_q =
    \bigl|
        \{
            \operatorname{id}(z):
            z\text{ is a model response in the Native trajectory for }q
        \}
    \bigr|.
    \label{eq:diag-horizon}
\end{equation}
The extractor counts unique identifiers with \code{object == "response"}.
$H_q$ measures observed execution length, not intrinsic task difficulty.
Native outcome labels are not used for grouping or performance comparison.
We sort by $(H_q,\mathrm{task\_id})$ and form three near-equal groups,
keeping each task's group fixed across metrics.
Table~\ref{tab:horizon-groups} gives the ranges and group sizes.

\begin{table}[!htbp]
\centering
\small
\setlength{\tabcolsep}{6pt}
\caption{Execution-length groups defined from independent Native response counts.}
\label{tab:horizon-groups}
\begin{tabular}{@{}lrrr@{}}
\toprule
\textbf{Group}
& \textbf{Native responses}
& \textbf{Median}
& \textbf{Tasks} \\
\midrule
Short  & 10--197  & 145 & 37 \\
Medium & 198--290 & 248 & 38 \\
Long   & 291--526 & 343 & 38 \\
\bottomrule
\end{tabular}
\end{table}

\begin{figure}[!t]
    \centering
    \includegraphics[width=\linewidth]
        {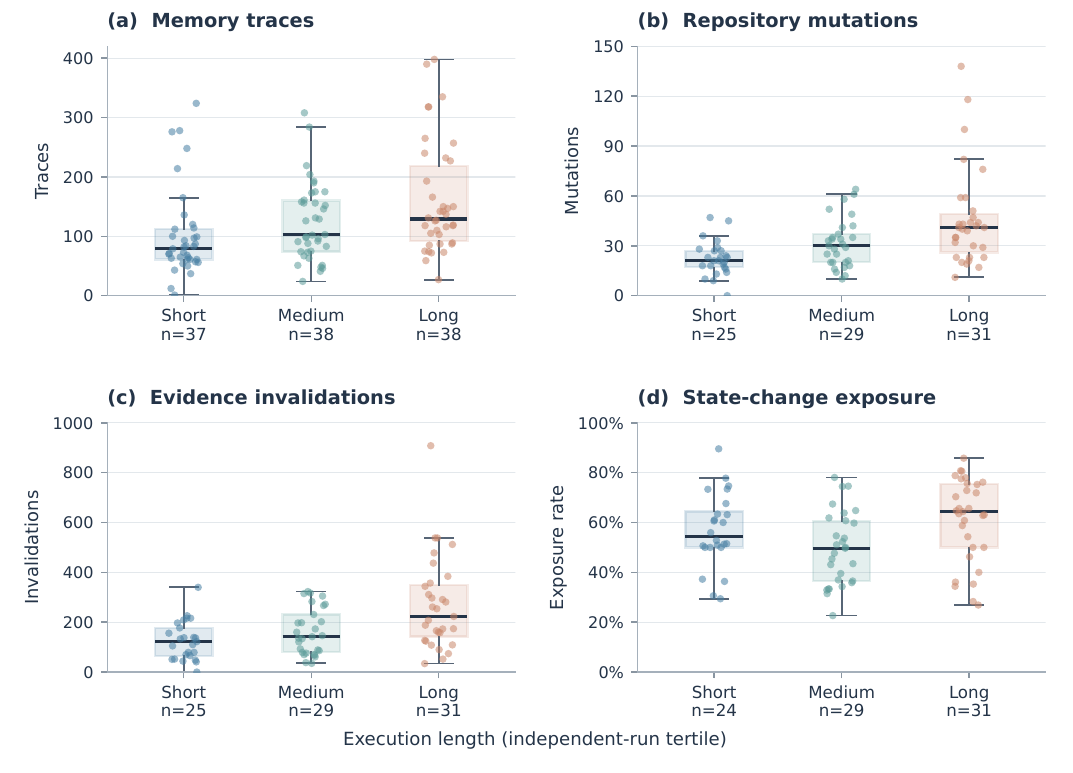}
    \caption{Memory behavior across short-, medium-, and long-horizon tasks.
    (a) Stored Memory Traces.
    (b) Repository revisions.
    (c) Stored observations or facts closed by MemTrace's revision policy
    (semantic evidence invalidations; Appendix~\ref{app:state-analysis}).
    (d) Fraction of file-bound Memory Traces whose bound files are later modified.
    Points are tasks; boxes show medians and interquartile ranges.
    Whiskers extend to observations within 1.5 interquartile ranges.
    Group sizes are 37/38/38 in (a), 25/29/31 in (b--c), and 24/29/31 in (d).
    Groups are defined by response counts from an independent Native run.}
    \label{fig:long-horizon-behavior}
\end{figure}

\paragraph{Measures and distributions.}
Stored Memory Traces are counted from their manifests.
Revisions $M_q$, invalidations $I_q$, and exposure $E_q$ follow
Appendix~\ref{app:state-analysis}.
Context Refreshes count
\codepath{NEW_EPOCH_MODEL_OBSERVED_AND_ACTIVATED} events.
The 258 events across 85 trajectories match the recorded epoch counts.
Figure~\ref{fig:long-horizon-behavior} shows task-level distributions;
Table~\ref{tab:horizon-descriptives} gives medians and interquartile ranges.

\begin{table}[!htbp]
\centering
\small
\setlength{\tabcolsep}{4pt}
\caption{Memory behavior by execution-length group: median [25th, 75th percentile]. Group-specific sample sizes are given in Figure~\ref{fig:long-horizon-behavior}; Context Refresh counts use the same tasks as revision counts.}
\label{tab:horizon-descriptives}
\begin{tabular}{@{}lccc@{}}
\toprule
\textbf{Metric}
& \textbf{Short}
& \textbf{Medium}
& \textbf{Long} \\
\midrule
Stored Memory Traces
& 79 [61,112]
& 102.5 [74.25,160.25]
& 129 [92.5,218.5] \\
Repository revisions
& 21 [17,27]
& 30 [20,37]
& 41 [26,49] \\
Evidence invalidations
& 122 [66,177]
& 142 [79,231]
& 223 [142.5,350.5] \\
Context Refreshes
& 2 [1,2]
& 2 [2,4]
& 3 [2,4.5] \\
Exposure (\%)
& 54.4 [50.0,64.4]
& 49.7 [36.5,60.7]
& 64.4 [50.0,75.5] \\
\bottomrule
\end{tabular}
\end{table}

\paragraph{Association with execution length.}
From Short to Long, median Memory Trace counts rise from 79 to 129,
revisions from 21 to 41, and evidence invalidations from 122 to 223.
Their correlations with Native response count are
$\rho=0.385$, $0.466$, and $0.403$, respectively
(Table~\ref{tab:horizon-associations}).
Context Refresh count is also positively associated with response count.
Exposure has no clear monotonic association
($\rho=0.137$; 95\% CI: $[-0.098,0.361]$).
Longer measured horizons are therefore associated with more stored history
and revision-triggered maintenance, rather than a consistently higher
exposure fraction.

\begin{table}[!htbp]
\centering
\small
\setlength{\tabcolsep}{4pt}
\caption{Association with Native response count and the Long--Short median difference. Intervals are 95\% task-bootstrap intervals. Exposure differences are percentage points; all other differences are counts.}
\label{tab:horizon-associations}
\begin{tabular}{@{}lrcc@{}}
\toprule
\textbf{Metric}
& $N$
& \textbf{Spearman $\rho$ [95\% CI]}
& \textbf{Long--Short [95\% CI]} \\
\midrule
Stored Memory Traces
& 113
& $0.385\ [0.206,0.539]$
& $50\ [27,78]$ \\
Repository revisions
& 85
& $0.466\ [0.285,0.614]$
& $20\ [9,25]$ \\
Evidence invalidations
& 85
& $0.403\ [0.205,0.566]$
& $101\ [33,206]$ \\
Context Refreshes
& 85
& $0.382\ [0.171,0.560]$
& $1\ [0,2]$ \\
Exposure (pp)
& 84
& $0.137\ [-0.098,0.361]$
& $10.0\ [-1.5,19.2]$ \\
\bottomrule
\end{tabular}
\end{table}

\Needspace{5\baselineskip}
\paragraph{Alternative execution-length measures.}
Native response count correlates with tool calls ($\rho=0.980$),
wall-clock time ($\rho=0.677$), and output tokens ($\rho=0.508$).
Using tool calls or output tokens for grouping preserves 97/113 (85.8\%)
or 52/113 (46.0\%) of the response-based assignments, respectively.
Memory Trace counts and invalidations are positively associated with
all three length measures (Table~\ref{tab:horizon-sensitivity}).
Revisions and Context Refreshes show stronger associations with responses
and tool calls than with output tokens.

\begin{table}[!htbp]
\centering
\footnotesize
\setlength{\tabcolsep}{3pt}
\caption{Spearman correlations under alternative Native execution-length measures, with 95\% task-bootstrap intervals. Sample sizes follow Table~\ref{tab:horizon-associations}.}
\label{tab:horizon-sensitivity}
\begin{tabular}{@{}lccc@{}}
\toprule
\textbf{Metric}
& \textbf{Responses}
& \textbf{Tool calls}
& \textbf{Output tokens} \\
\midrule
Stored Memory Traces
& $0.385\ [0.206,0.539]$
& $0.384\ [0.209,0.538]$
& $0.200\ [0.014,0.367]$ \\
Repository revisions
& $0.466\ [0.285,0.614]$
& $0.463\ [0.278,0.616]$
& $0.199\ [-0.002,0.380]$ \\
Evidence invalidations
& $0.403\ [0.205,0.566]$
& $0.401\ [0.201,0.570]$
& $0.256\ [0.052,0.440]$ \\
Context Refreshes
& $0.382\ [0.171,0.560]$
& $0.385\ [0.175,0.564]$
& $0.187\ [-0.039,0.392]$ \\
Exposure
& $0.137\ [-0.098,0.361]$
& $0.126\ [-0.106,0.348]$
& $0.214\ [-0.011,0.432]$ \\
\bottomrule
\end{tabular}
\end{table}

\begin{figure}[!htbp]
    \centering
    \includegraphics[width=\linewidth]
        {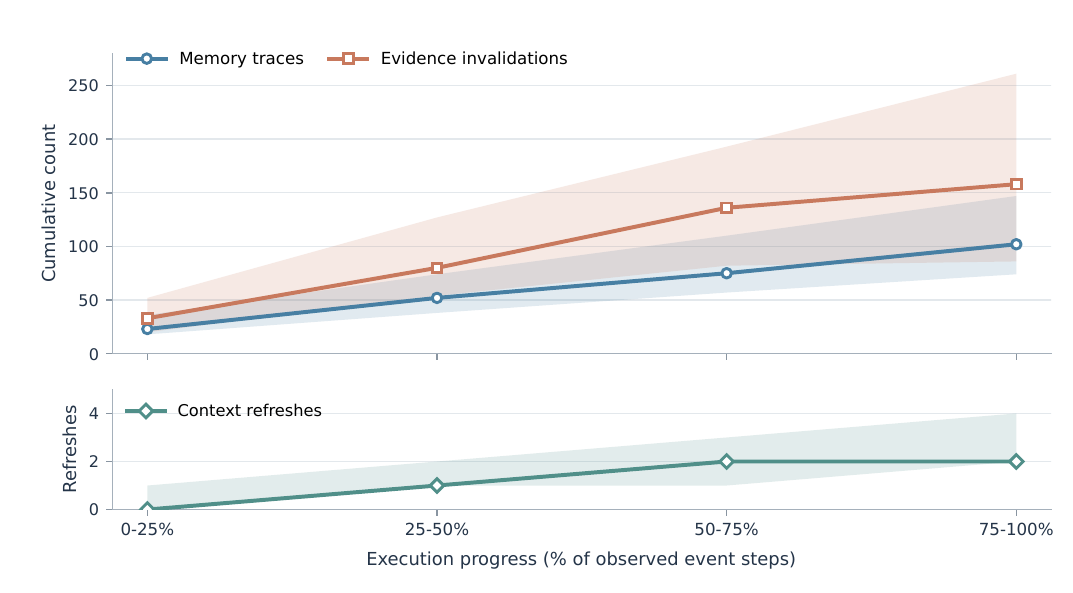}
    \caption{Cumulative counts across execution progress in 85 instrumented trajectories.
    Lines show task medians and shaded regions show interquartile ranges.
    Progress uses internal event steps.
    The upper panel shows Memory Traces and evidence invalidations;
    the lower panel shows Context Refreshes on a separate scale.}
    \label{fig:execution-progress}
\end{figure}

\paragraph{Within-run accumulation.}
We split each instrumented trajectory into four equal event-step intervals
and count traces and events accumulated by each interval endpoint.
Median cumulative Memory Trace counts are 23, 52, 75, and 102;
median cumulative invalidations are 33, 80, 136, and 158.
Figure~\ref{fig:execution-progress} separates Context Refreshes from
the larger trace and invalidation counts.

\Needspace{5\baselineskip}
\subsection{Case Study: Repository Evolution Across a Context Refresh}
\label{app:case-study}

\paragraph{Task and selection.}
We examine the successful archived mini-swe-agent trajectory for
\codepath{scriggo-method-declarations}.
The task adds method declarations to Scriggo, including value and pointer
receivers, method expressions, and interface dispatch~\citep{huang2026deepswe}.
Its records link a sealed Memory Trace to a later bound-file revision,
a Context Refresh, and subsequent verification.

\paragraph{Trace and repository evolution.}
At event step 1,061, Memory Trace $\tau_{\mathrm{A}}$
is sealed with a binding to
\codepath{internal/compiler/parser_test.go}.
At step 17,152, that file is modified in a revision recording
15 semantic evidence invalidations.
The shared file path and event order establish exposure for
$\tau_{\mathrm{A}}$ under Eq.~\eqref{eq:diag-exposure-indicator}.
The invalidation count applies to the revision, not specifically to this trace.

\begin{figure}[!htbp]
    \centering
    \includegraphics[width=\linewidth]{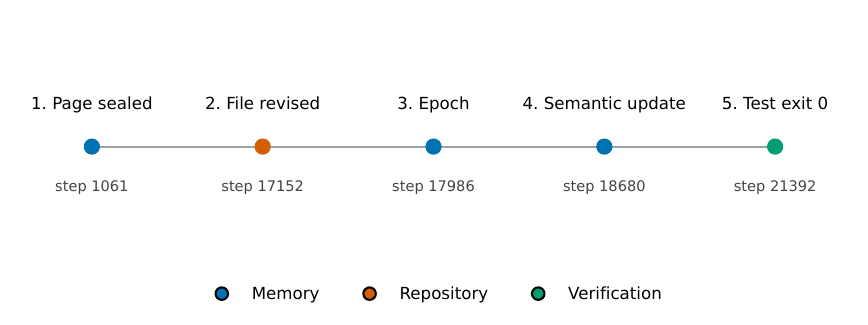}
    \caption{Selected events from the Scriggo trajectory.
    Events show Memory Trace sealing, a bound-file revision, Context Refresh,
    a semantic update, and successful in-run verification.
    Positions show event order, not elapsed-time spacing.}
    \label{fig:case-study}
\end{figure}

\paragraph{Continuation and verification.}
A new context epoch is activated at step 17,986.
At step 18,680, two semantic updates are accepted for
Execution Milestone \code{M003}.
At step 21,392, \code{go test ./compare/ -count=1} returns exit code 0
for criterion \code{M003.C001}.
The separate archived score summary records Harbor reward 1.
Table~\ref{tab:case-events} summarizes the logged sequence.

\begin{table}[!htbp]
\centering
\small
\setlength{\tabcolsep}{4pt}
\renewcommand{\arraystretch}{1.12}
\caption{Recorded events in the Scriggo case. File bindings come from the Memory Trace manifest; step numbers use the internal event clock.}
\label{tab:case-events}
\begin{tabular}{@{}rL{0.25\linewidth}L{0.60\linewidth}@{}}
\toprule
\textbf{Step} & \textbf{Record} & \textbf{Observed information} \\
\midrule
1,061 & Memory Trace sealing
& $\tau_{\mathrm{A}}$ is sealed; its manifest includes the compiler parser-test file. \\
17,152 & Repository revision
& The bound file is modified; the revision records 15 evidence invalidations. \\
17,986 & Context Refresh
& A new context epoch is activated after model observation. \\
18,680 & Semantic update
& Two semantic updates are accepted for \code{M003}. \\
21,392 & Runtime verification
& The comparison test command exits with code 0; criterion \code{M003.C001} is marked successful. \\
\bottomrule
\end{tabular}
\end{table}

\paragraph{Interpretation.}
The case links historical storage, a later change to a bound file,
and successful continuation across a Context Refresh.
The file binding identifies where current applicability must be checked;
the later test supplies evidence about the updated execution state.
This illustrates the distinction between preserving history and
establishing its current applicability.

\end{document}